\documentclass[10pt,twocolumn,letterpaper]{article}

\usepackage{cvpr}
\usepackage{times}
\usepackage{epsfig}
\usepackage{graphicx}
\usepackage{amsmath}
\usepackage{amssymb}
\usepackage{xfrac}
\usepackage{url}
\usepackage{color}
\usepackage{caption}
\usepackage{booktabs}
\usepackage{multirow}
\usepackage{ctable}

\usepackage[breaklinks=true,bookmarks=false]{hyperref}

\cvprfinalcopy 

\begin{document}

\title{Recurrent MVSNet for High-resolution Multi-view Stereo Depth Inference}

\author{
Yao Yao$^1$\thanks{Yao Yao is an intern at Shenzhen Zhuke Innovation Technology.}
\and
Zixin Luo$^1$
\and
Shiwei Li$^1$
\and
Tianwei Shen$^1$
\and
Tian Fang$^2$
\and
Long Quan$^1$
\vspace{1mm}
\and
\centerline{$^1$The Hong Kong University of Science and Technology} \\
\centerline{\tt\small $\{$yyaoag, zluoag, slibc, tshenaa, quan$\}$@cse.ust.hk} \\
\centerline{$^2$Shenzhen Zhuke Innovation Technology (Altizure)} \\
\centerline{\tt\small fangtian@altizure.com}
}

\maketitle

\begin{abstract}

Deep learning has recently demonstrated its excellent performance for multi-view stereo (MVS). However, one major limitation of current learned MVS approaches is the scalability: the memory-consuming cost volume regularization makes the learned MVS hard to be applied to high-resolution scenes. 
In this paper, we introduce a scalable multi-view stereo framework based on the recurrent neural network. Instead of regularizing the entire 3D cost volume in one go, the proposed Recurrent Multi-view Stereo Network (R-MVSNet) sequentially regularizes the 2D cost maps along the depth direction via the gated recurrent unit (GRU).
This reduces dramatically the memory consumption and makes high-resolution reconstruction feasible. 
We first show the state-of-the-art performance achieved by the proposed R-MVSNet on the recent MVS benchmarks. Then, we further demonstrate the scalability of the proposed method on several large-scale scenarios, where previous learned approaches often fail due to the memory constraint. Code is available at \url{https://github.com/YoYo000/MVSNet}.

\end{abstract}

\section{Introduction}
\noindent
Multi-view stereo (MVS) aims to recover the dense representation of the scene given multi-view images and calibrated cameras. While traditional methods have achieved excellent performance on MVS benchmarks, recent works \cite{ji2017surfacenet,huang2018deepmvs,yao2018mvsnet} show that learned approaches are able to produce results comparable to the traditional state-of-the-arts. In particular, MVSNet \cite{yao2018mvsnet} proposed a deep architecture for depth map estimation, which significantly boosts the reconstruction completeness and the overall quality. 

\begin{figure*}[t!]
  \centering
  \includegraphics[width=1\linewidth]{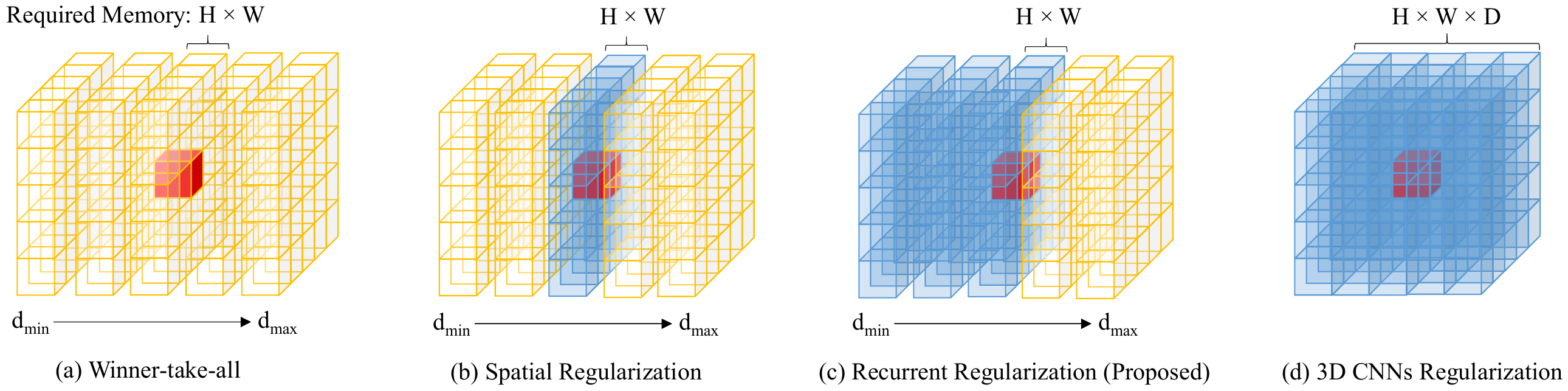}
  \caption{Illustrations of different regularization schemes. For the interest {\color{red}red} voxel, we use voxels in {\color{blue}blue} to denote its receptive field during the cost volume regularization. The runtime memory requirement is also listed on top of the volume, where H, W and D denote the image height, width and depth sample number respectively. The 3D CNNs gather the cost information across the whole space, however, requires a runtime memory cubical to the model resolution}
  \label{fig:sequential}
\end{figure*}

One of the key advantages of learning-based MVS is the cost volume regularization, where most networks apply multi-scale 3D CNNs \cite{ji2017surfacenet,kar2017learning,yao2018mvsnet} to regularize the 3D cost volume. However, this step is extremely memory expensive: it operates on 3D volumes and the memory requirement grows cubically with the model resolution (Fig.~\ref{fig:sequential} (d)). Consequently, current learned MVS algorithms could hardly be scaled up to high-resolution scenarios. 

Recent works on 3D with deep learning also acknowledge this problem. OctNet \cite{riegler2017octnet} and O-CNN \cite{wang2017cnn} exploit the sparsity in 3D data and introduce the octree structure to 3D CNNs. SurfaceNet \cite{ji2017surfacenet} and DeepMVS \cite{huang2018deepmvs} apply the engineered divide-and-conquer strategy to the MVS reconstruction. MVSNet \cite{yao2018mvsnet} builds the cost volume upon the reference camera frustum to decouple the reconstruction into smaller problems of per-view depth map estimation. However, when it comes to a high-resolution 3D reconstruction (e.g., volume size $> 512^3$ voxels), these methods will either fail or take a long time for processing.  

To this end, we present a novel scalable multi-view stereo framework, dubbed as R-MVSNet, based on the recurrent neural network. The proposed network is built upon the MVSNet architecture \cite{yao2018mvsnet}, but regularizes the cost volume in a sequential manner using the convolutional gated recurrent unit (GRU) rather than 3D CNNs. With the sequential processing, the online memory requirement of the algorithm is reduced from cubic to quadratic to the model resolution (Fig.~\ref{fig:sequential} (c)). As a result, the R-MVSNet is applicable to high resolution 3D reconstruction with \textit{unlimited} depth-wise resolution. 

We first evaluate the R-MVSNet on DTU \cite{aanaes2016large}, Tanks and Temples \cite{knapitsch2017tanks} and ETH3D \cite{schoeps2017cvpr} datasets, where our method produces results comparable or even outperforms the state-of-the-art MVSNet \cite{yao2018mvsnet}. 
Next, we demonstrate the scalability of the proposed method on several large-scale scenarios with detailed analysis on the memory consumption. R-MVSNet is much more efficient than other methods in GPU memory and is the first learning-based approach applicable to such wide depth range scenes, e.g., the \textit{advance} set of Tanks and Temples dataset \cite{knapitsch2017tanks}.

\section{Related Work}

\paragraph{Learning-based MVS Reconstruction}
Recent learning-based approaches have shown great potentials for MVS reconstruction. Multi-patch similarity \cite{hartmann2017learned} is proposed to replace the traditional cost metric with the learned one. SurfaceNet \cite{ji2017surfacenet} and DeepMVS \cite{huang2018deepmvs} pre-warp the multi-view images to 3D space, and regularize the cost volume using CNNs. LSM \cite{kar2017learning} proposes differentiable projection operations to enable the end-to-end MVS training. Our approach is mostly related to MVSNet \cite{yao2018mvsnet}, which encodes camera geometries in the network as differentiable homography and infers the depth map for the reference image. While some methods have achieved excellent performance in MVS benchmarks, aforementioned learning-based pipelines are restricted to small-scale MVS reconstructions due to the memory constraint. 

\paragraph{Scalable MVS Reconstruction} 
The memory requirement of learned cost volume regularizations \cite{ji2017surfacenet,kar2017learning,huang2018deepmvs,chang2018pyramid,yao2018mvsnet} grows cubically with the model resolution, which will be intractable when large image sizes or wide depth ranges occur. Similar problem also exists in traditional MVS reconstructions (e.g., semi-global matching \cite{hirschmuller2008stereo}) if the whole volume is taken as the input to the regularization. To mitigate the scalability issue, learning-based OctNet \cite{riegler2017octnet} and O-CNN \cite{wang2017cnn} exploit the sparsity in 3D data and introduce the octree structure to 3D CNNs, but are still restricted to reconstructions with resolution $< 512^3$ voxels. Heuristic divide-and-conquer strategies are applied in both classical \cite{kuhn2017tv} and learned MVS approaches \cite{ji2017surfacenet,huang2018deepmvs}, however, usually lead to the loss of global context information and the slow processing speed. 

On the other hand, scalable traditional MVS algorithms all regularize the cost volume implicitly. They either apply local depth propagation \cite{lhuillier2005quasi,furukawa2010accurate,galliani2015massively,schonberger2016pixelwise} to iteratively refine depth maps/point clouds, or sequentially regularize the cost volume using simple plane sweeping \cite{collins1996space} and 2D spatial cost aggregation with depth-wise winner-take-all \cite{yang2012non,yoon2006adaptive}. In this work, we follow the idea of sequential processing, and propose to regularize the cost volume using the convolutional GRU \cite{cho2014learning}. GRU is a RNN architecture \cite{elman1990finding} initially proposed for learning sequential speech and text data, and is recently applied to 3D volume processing, e.g., video sequence analysis \cite{ballas2015delving,zhu2018towards}. For our task, the convolutional GRU gathers spatial as well as temporal context information in the depth direction, which is able to achieve comparable regularization results to 3D CNNs.


\begin{figure*}
  \centering
  \includegraphics[width=1\linewidth]{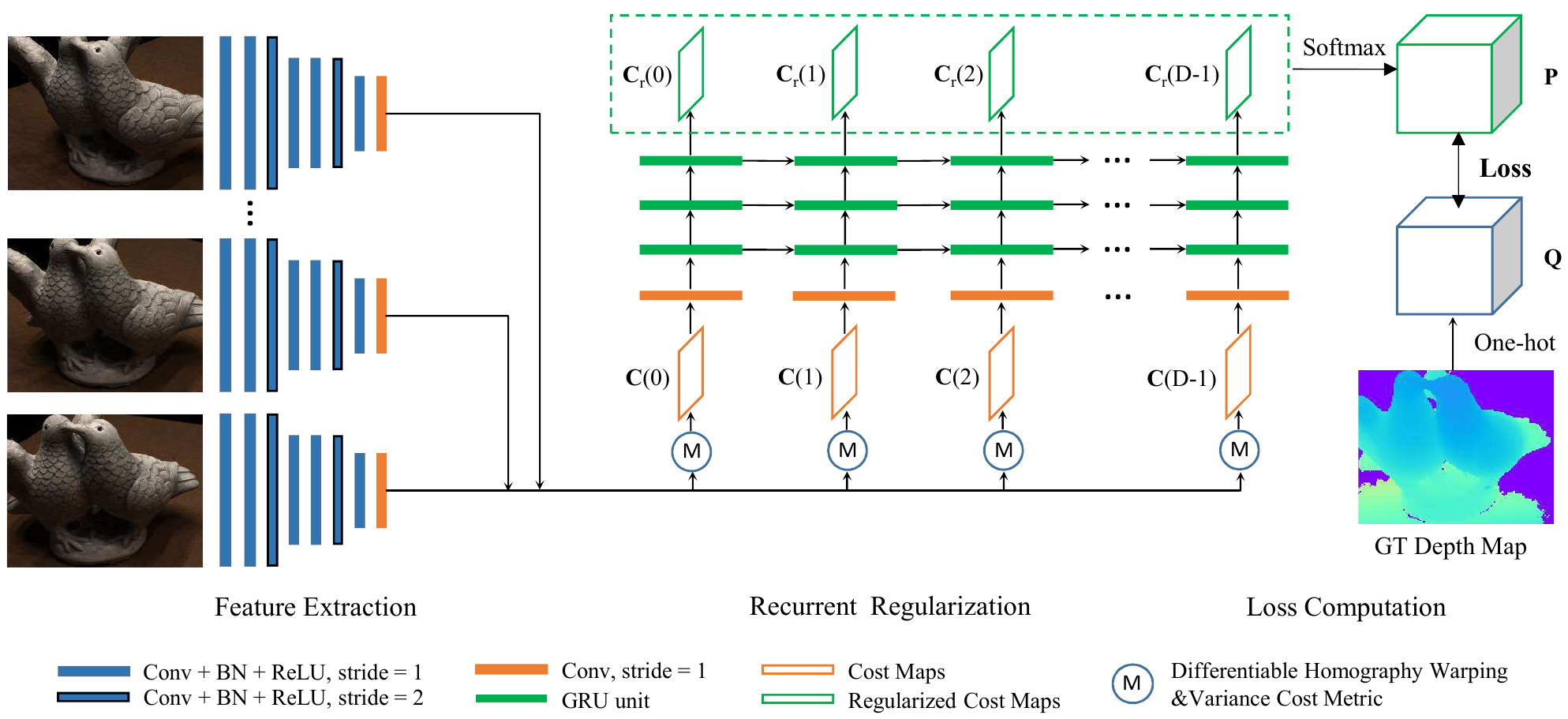}
  \caption{The R-MVSNet architecture. Deep image features are extracted from input images and then warped to the fronto-parallel planes of the reference camera frustum. The cost maps are computed at different depths and are sequentially regularized by the convolutional GRU. The network is trained as a classification problem with the cross-entropy loss}
  \label{fig:r-network}
\end{figure*}

\section{Network Architecture}
\noindent
This section describes the detailed network architecture of R-MVSNet. Our method can be viewed as an extension to the recent MVSNet \cite{yao2018mvsnet} with cost volume regularization using convolutional GRU. We first review the MVSNet architecture in Sec.~\ref{sec:mvsnet}, and then introduce the recurrent regularization in Sec.~\ref{recurrent} and the corresponding loss formulation in Sec.~\ref{loss}.

\subsection{Review of MVSNet} \label{sec:mvsnet}
\noindent
Given a reference image $\mathbf{I}_1$ and a set of its neighboring source images $\{\mathbf{I}_i\}_{i=2}^N$, MVSNet \cite{yao2018mvsnet} proposes an end-to-end deep neural network to infer the reference depth map $\mathbf{D}$. In its network, deep image features $\{\mathbf{F}_i\}_{i=1}^N$ are first extracted from input images through a 2D network. These 2D image features will then be warped into the reference camera frustum by differentiable homographies to build the feature volumes $\{\mathbf{V}_i\}_{i=1}^N$ in 3D space. To handle arbitrary $N$-view image input, a variance based cost metric is proposed to map N feature volumes to one cost volume $\mathbf{C}$. Similar to other stereo and MVS algorithms, MVSNet regularizes the cost volume using the multi-scale 3D CNNs, and regresses the reference depth map $\mathbf{D}$ through the soft argmin \cite{kendall2017end} operation. A refinement network is applied at the end of MVSNet to further enhance the depth map quality. As deep image features $\{\mathbf{F}_i\}_{i=1}^N$ are downsized during the feature extraction, the output depth map size is $1 / 4$ to the original image size in each dimension.

MVSNet has shown state-of-the-art performance on DTU dataset \cite{aanaes2016large} and the \textit{intermediate} set of Tanks and Temples dataset \cite{knapitsch2017tanks}, which contain scenes with outside-looking-in camera trajectories and small depth ranges. However, MVSNet can only handle a maximum reconstruction scale at $H \times W \times D = 1600 \times 1184 \times 256$ with the 16 GB large memory Tesla P100 GPU, and will fail at larger scenes e.g., the \textit{advanced} set of Tanks and Temples. To resolve the scalability issue especially for the wide depth range reconstructions, we will introduce the novel recurrent cost volume regularization in the next section.

\subsection{Recurrent Regularization} \label{recurrent}

\paragraph{Sequential Processing} 
An alternative to globally regularize the cost volume $\mathbf{C}$ in one go is to sequentially process the volume through the depth direction. 
The simplest sequential approach is the winner-take-all plane sweeping stereo~\cite{collins1996space}, which crudely replaces the pixel-wise depth value with the better one and thus suffers from noise (Fig.~\ref{fig:sequential} (a)). To improve, cost aggregation methods \cite{yang2012non,yoon2006adaptive} filter the matching cost $\mathbf{C}(d)$ at different depths (Fig.~\ref{fig:sequential} (b)) so as to gather spatial context information for each cost estimation.  
In this work, we follow the idea of sequential processing, and propose a more powerful recurrent regularization scheme based on convolutional GRU. The proposed method is able to gather spatial as well as the uni-directional context information in the depth direction (Fig.~\ref{fig:sequential} (c)), which achieves regularization results comparable to the full-space 3D CNNs but is much more efficient in runtime memory. 


\begin{figure*}
  \centering
  \includegraphics[width=1\linewidth]{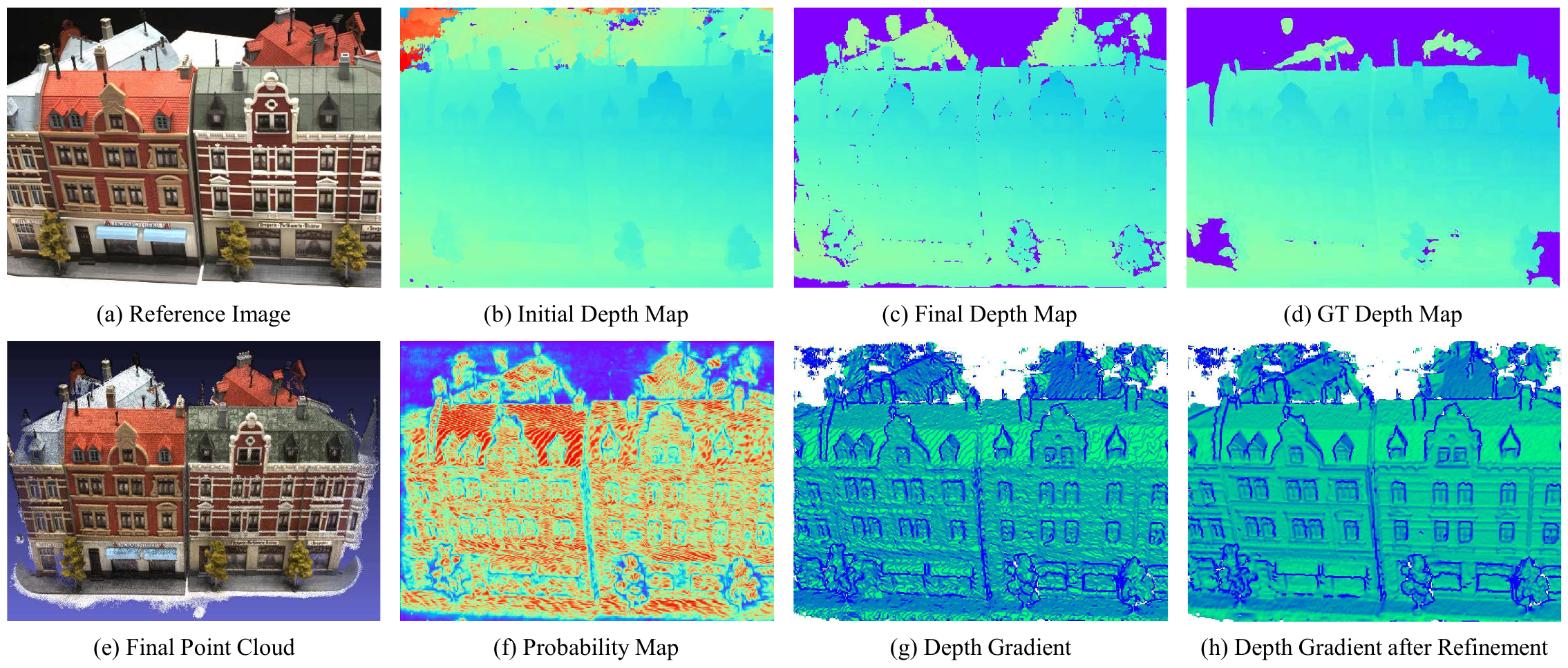}
  \caption{Reconstruction pipeline. (a) Image 24 of DTU \cite{aanaes2016large} \textit{scan 15}. (b) Initial depth map from the network. (c) Final depth map (Sec.~\ref{sec:post}). (d) Ground truth depth map. (e) Point cloud output. (f) Probability estimation for depth map filtering (Sec.~\ref{sec:post}). (g) The gradient visualization of the initial depth map. (h) The gradient visualization after the refinement (Sec.~\ref{sec:refinement})}
  \label{fig:pipeline}
\end{figure*}

\paragraph{Convolutional GRU}
Cost volume $\mathbf{C}$ could be viewed as $D$ cost maps $\{\mathbf{C}(i)\}_{i=1}^{D}$ concatenated in the depth direction. If we denote the output of regularized cost maps as $\{\mathbf{C}_r(i)\}_{i=1}^{D}$, for the ideal sequential processing at the $t^{th}$ step, $\mathbf{C}_r(t)$ should be dependent on cost maps of the current step $\mathbf{C}(t)$ as well as all previous steps $\{\mathbf{C}(i)\}_{i=1}^{t-1}$. Specifically, in our network we apply a convolutional variant of GRU to aggregate such temporal context information in depth direction, which corresponds to the time direction in language processing. In the following, we denote `$\odot$' as the element-wise multiplication, `[]' the concatenation and `$*$' the convolution operation. Cost dependencies are formulated as:
\begin{equation}
\mathbf{C}_r(t) = (1 - \mathbf{U}(t)) \odot \mathbf{C}_r(t - 1) + \mathbf{U}(t) \odot \mathbf{C}_u(t)
\end{equation}
where $\mathbf{U}(t)$ is the update gate map to decide whether to update the output for current step, $\mathbf{C}_r(t - 1)$ is the regularized cost map of late step, and $\mathbf{C}_u(t)$ could be viewed as the updated cost map in current step, which is defined as:
\begin{equation}
\mathbf{C}_u(t) = \mathcal{\sigma}_c(\mathbf{W}_c * [\mathbf{C}(t), \mathbf{R}(t) \odot \mathbf{C}_r(t - 1)] + \mathbf{b}_c)
\end{equation}
$\mathbf{R}(t)$ here is the reset gate map to decide how much the previous $\mathbf{C}_r(t - 1)$ should affect the current update. $\sigma_c(\cdot)$ is the nonlinear mapping, which is the element-wise sigmoid function. The update gate and reset gate maps are also related to the current input and previous output:
\begin{equation}
\mathbf{R}(t) = \mathcal{\sigma}_g(\mathbf{W}_r * [\mathbf{C}(t), \mathbf{C}_r(t - 1)] + \mathbf{b}_r)
\end{equation}
\begin{equation}
\mathbf{U}(t) = \mathcal{\sigma}_g(\mathbf{W}_u * [\mathbf{C}(t), \mathbf{C}_r(t - 1)] + \mathbf{b}_u)
\end{equation}
$\mathbf{W}$ and $\mathbf{b}$ are learned parameters. The nonlinear $\sigma_g(\cdot)$ is the hyperbolic tangent to make soft decisions for the updates. 

The convolutional GRU architecture not only spatially regularizes the cost maps through 2D convolutions, but also aggregates the temporal context information in depth direction. We will show in the experiment section that our GRU regularization can significantly outperform the simple winner-take-all or only the spatial cost aggregation.

\paragraph{Stacked GRU} 
The basic GRU model is comprised of a single layer. To further enhance the regularization ability, more GRU units could be stacked to make a deeper network. In our experiments, we adopt a 3-layer stacked GRU structure (Fig.~\ref{fig:r-network}). Specifically, we first apply a 2D convolutional layer to map the 32-channel cost map $\mathbf{C}(t)$ to 16-channel as the input to the first GRU layer. The output of each GRU layer will be used as the input to the next GRU layer, and the output channel numbers of the 3 layers are set to 16, 4, 1 respectively. The regularized cost maps $\{\mathbf{C}_r(i)\}_{i=1}^{D}$ will finally go through a softmax layer to generate the probability volume $\mathbf{P}$ for calculating the training loss.

\subsection{Training Loss} \label{loss}
\noindent
Most deep stereo/MVS networks regress the disparity/depth outputs using the \textit{soft argmin} operation \cite{kendall2017end}, which can be interpreted as the expectation value along the depth direction \cite{yao2018mvsnet}. The expectation formulation is valid if depth values are \textit{uniformly} sampled within the depth range. However, in recurrent MVSNet, we apply the \textit{inverse depth} to sample the depth values in order to efficiently handle reconstructions with wide depth ranges. Rather than treat the problem as a regression task, we train the network as a multi-class classification problem with cross entropy loss:
\begin{equation}
Loss = \sum\limits_{\mathbf{p}} \Big( \sum\limits_{i = 1}^{D} -\mathbf{P}(i, \mathbf{p}) \cdot \log \mathbf{Q}(i, \mathbf{p}) \Big)
\end{equation}
where $\mathbf{p}$ is the spatial image coordinate and $\mathbf{P}(i, \mathbf{p})$ is a voxel in the probability volume $\mathbf{P}$. $\mathbf{Q}$ is the ground truth binary occupancy volume, which is generated by the one-hot encoding of the ground truth depth map. $\mathbf{Q}(i, \mathbf{p})$ is the corresponding voxel to $\mathbf{P}(i, \mathbf{p})$. 

One concern about the classification formulation is the discretized depth map output \cite{zbontar2016stereo,luo2016efficient,huang2018deepmvs}. To achieve sub-pixel accuracy, a variational depth map refinement algorithm is proposed in Sec.~\ref{sec:refinement} to further refine the depth map output. In addition, while we need to compute the whole probability volume during training, for testing, the depth map can be sequentially retrieved from the regularized cost maps using the winner-take-all selection.

\section{Reconstruction Pipeline}
\noindent
The proposed network in the previous section generates the depth map per-view. This section describes the non-learning parts of our 3D reconstruction pipeline.

\subsection{Preprocessing}\label{sec:pre}
\noindent
To estimate the reference depth map using R-MVSNet, we need to prepare: 1) the source images $\{\mathbf{I}_i\}_{i=2}^N$ of the given reference image $\mathbf{I}_1$, 2) the depth range $[d_{min}, d_{max}]$ of the reference view and 3) the depth sample number $D$ for sampling depth values using the \textit{inverse depth} setting. 

For selecting the source images, we follow MVSNet \cite{yao2018mvsnet} to score each image pair using a piece-wise Gaussian function w.r.t. the baseline angle of the sparse point cloud \cite{zhang2015joint}. The neighboring source images are selected according to the pair scores in descending order. The depth range is also determined by the sparse point cloud with the implementation of COLMAP \cite{schonberger2016pixelwise}. Depth samples are chosen within $[d_{min}, d_{max}]$ using the inverse depth setting and we determine the total depth sample number $D$ by adjusting the temporal depth resolution to the spatial image resolution (details are described in the supplementary material).

\subsection{Variational Depth Map Refinement} \label{sec:refinement}
\noindent
As mentioned in Sec.~\ref{loss}, a depth map will be retrieved from the regularized cost maps through the winner-take-all selection. Compare to the \textit{soft argmin} \cite{kendall2017end} operation, the \textit{argmax} operation of winner-take-all cannot produce depth estimations with sub-pixel accuracy. To alleviate the stair effect (see Fig.~\ref{fig:pipeline} (g) and (h)), we propose to refine the depth map in a small depth range by enforcing the multi-view photo-consistency.

Given the reference image $\mathbf{I}_1$, the reference depth map $\mathbf{D}_1$ and one source image $\mathbf{I}_i$, we project $\mathbf{I}_i$ to $\mathbf{I}_1$ through $\mathbf{D}_1$ to form the reprojected image $\mathbf{I}_{i \to 1}$. The image reprojection error between $\mathbf{I}_1$ and $\mathbf{I}_{i \to 1}$ at pixel $\mathbf{p}$ is defined as:
\begin{equation}
\begin{split}
E^{i}(\mathbf{p}) &= E^i_{photo}(\mathbf{p}) + E^{i}_{smooth}(\mathbf{p})  \\
			  &= \mathcal{C}(\mathbf{I}_1(\mathbf{p}), \mathbf{I}_{i \to 1}(\mathbf{p})) +
				\sum_{\mathbf{p}' \in \mathcal{N}(\mathbf{p})}\mathcal{S}(\mathbf{p},\mathbf{p'})			
\end{split}
\end{equation}
where $E_{photo}^i$ is the photo-metric error between two pixels, $E_{smooth}^i$ is the regularization term to ensure the depth map smoothness. We choose the zero-mean normalized cross-correlation (ZNCC) to measure the photo-consistency $\mathcal{C}(\cdot)$, and use the bilateral squared depth difference $\mathcal{S}(\cdot)$ between $\mathbf{p}$ and its neighbors $\mathbf{p}' \in \mathcal{N}(\mathbf{p})$ for smoothness.

During the refinement, we iteratively minimize the total image reprojection error between the reference image and all source images $E = \sum_i \sum_{\mathbf{p}} E_{i \to 1}(\mathbf{p})$ w.r.t. depth map $\mathbf{D}_1$. It is noteworthy that the initial depth map from R-MVSNet has already achieved satisfying result. The proposed variational refinement only fine-tunes the depth values within a small range to achieve sub-pixel depth accuracy, which is similar to the quadratic interpolation in stereo methods \cite{zbontar2016stereo,luo2016efficient} and the DenseCRF in DeepMVS \cite{huang2018deepmvs}.

\subsection{Filtering and Fusion} \label{sec:post}
\noindent
Similar to other depth map based MVS approaches\cite{galliani2015massively,schonberger2016pixelwise,yao2018mvsnet}, we filter and fuse depth maps in R-MVSNet into a single 3D point cloud. The photo-metric and the geometric consistencies are considered in depth map filtering. As described in previous sections, the regularized cost maps will go through a softmax layer to generate the probability volume. In our experiments, we take the corresponding probability of the selected depth value as its confidence measurement (Fig.~\ref{fig:pipeline} (f)), and we will filter out pixels with probability lower than a threshold of $0.3$. The geometric constraint measures the depth consistency among multiple views, and we follow the geometric criteria in MVSNet \cite{yao2018mvsnet} that pixels should be at least three view visible. For depth map fusion, we apply the visibility-based depth map fusion \cite{merrell2007real} as well as the mean average fusion \cite{yao2018mvsnet} to further enhance the depth map quality and produce the 3D point cloud. Illustrations of our reconstruction pipeline are shown in Fig.~\ref{fig:pipeline}.

\section{Experiments}

\subsection{Implementation}

\paragraph{Training}
We train R-MVSNet on the DTU dataset \cite{aanaes2016large}, which contains over 100 scans taken under 7 different lighting conditions and fixed camera trajectories. While the dataset only provides the ground truth point clouds, we follow MVSNet \cite{yao2018mvsnet} to generate the rendered depth maps for training. The training image size is set to $W \times H = 640 \times 512$ and the input view number is $N=3$. The depth hypotheses are sampled from 425mm to 905mm with $D = 192$. In addition, to prevent depth maps from being biased on the GRU regularization order, each training sample is passed to the network with forward GRU regularization from $d_{min}$ to $d_{max}$ as well as the backward regularization from $d_{max}$ to $d_{min}$. The dataset is splitted into the same training, validation and evaluation sets as previous works \cite{ji2017surfacenet,yao2018mvsnet}. We choose TensorFlow \cite{tensorflow2015-whitepaper} for the network implementation, and the model is trained for 100k iterations with batch size of 1 on a GTX 1080Ti graphics card. RMSProp is chosen as the optimizer and the learning rate is set to 0.001 with an exponential decay of 0.9 for every 10k iterations.

\vspace{-4mm}
\paragraph{Testing}
For testing, we use $N=5$ images as input, and the inverse depth samples are adaptively selected as described in Sec.~\ref{sec:pre}. For Tanks and Temples dataset, the camera parameters are computed from OpenMVG \cite{openMVG} as suggested by MVSNet \cite{yao2018mvsnet}. Depth map refinement, filtering and fusion are implemented using OpenGL on the same GTX 1080Ti GPU.

\subsection{Benchmarks}\label{sec:benchmarks}
\noindent
We first demonstrate the state-of-the-art performance of the proposed R-MVSNet, which produces results comparable to or outperforms the previous MVSNet \cite{yao2018mvsnet}.

\vspace{-4mm}
\paragraph{DTU Dataset \cite{aanaes2016large}} 
We evaluate the proposed method on the DTU evaluation set. To compare R-MVSNet with MVSNet~\cite{yao2018mvsnet}, we set $[d_{min}, d_{max}] = [425, 905]$ and $D = 256$ for all scans. Quantitative results are shown in Table~\ref{tab:dtu}. The accuracy and the completeness are calculated using the matlab script provided by the DTU dataset. To summarize the overall reconstruction quality, we calculate the average of the mean accuracy and the mean completeness as the \textit{overall} score. Our R-MVSNet produces the best reconstruction completeness and overall score among all methods. Qualitative results can be found in Fig.~\ref{fig:dtu}.

\vspace{-4mm}
\paragraph{Tanks and Temples Benchmark \cite{knapitsch2017tanks}}
Unlike the indoor DTU dataset, Tanks and Temples is a large dataset captured in more complex environments. Specifically, the dataset is divided into the \textit{intermediate} and the \textit{advanced} sets. The \textit{intermediate} set contains scenes with outside-look-in camera trajectories, while the \textit{advanced} set contains large scenes with complex geometric layouts, where almost all previous learned algorithms fail due to the memory constraint. 

\begin{figure*}
  \centering
  \includegraphics[width=1\linewidth]{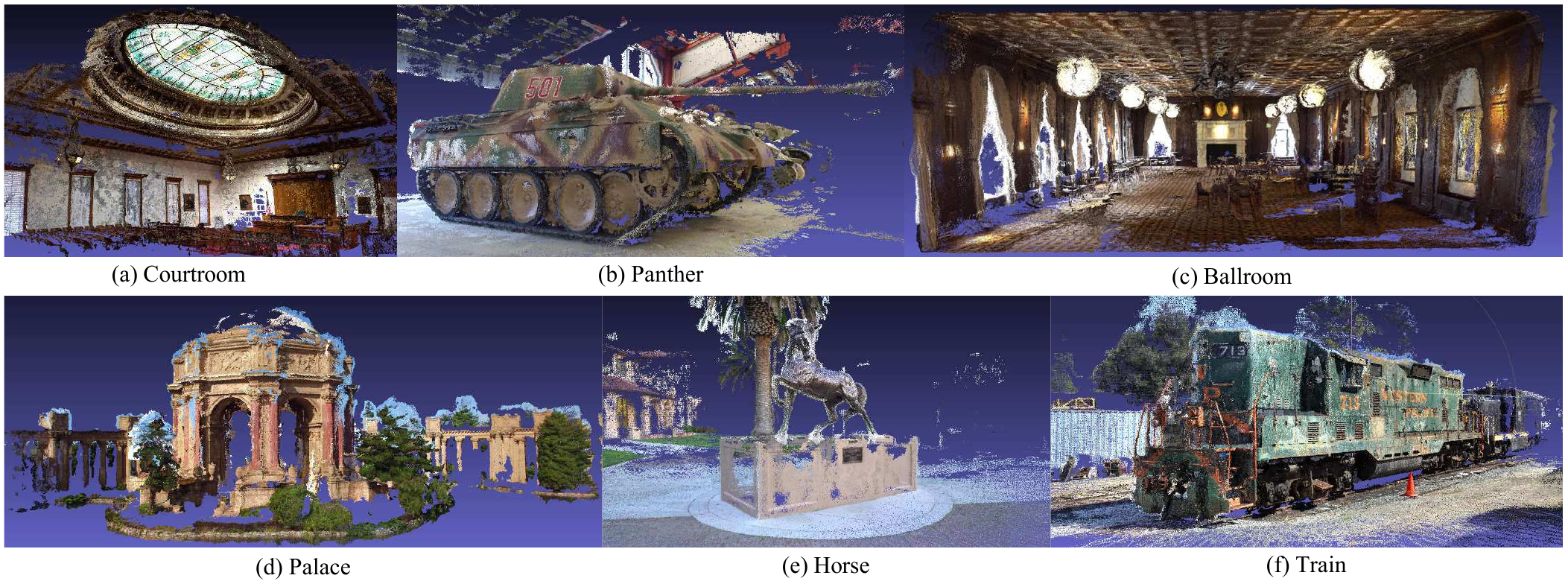}
	\vspace{-5mm}
  \caption{Point cloud reconstructions of Tanks and Temples dataset \cite{knapitsch2017tanks}}
  \label{fig:tanksandtemples}
\end{figure*}

The proposed method ranks $3^{rd}$ on the \textit{intermediate} set, which outperforms the original MVSNet \cite{yao2018mvsnet}. Moreover, R-MVSNet successfully reconstructs all scenes and also ranks $3^{rd}$ on the \textit{advanced} set. The reconstructed point clouds are shown in Fig.~\ref{fig:tanksandtemples}. It is noteworthy that the benchmarking result of Tanks and Temples is highly dependent on the point cloud density. Our depth map is of size $\frac{H}{4} \times \frac{W}{4}$, which is relatively low-resolution and will result in low reconstruction completeness. So for the evaluation, we linearly upsample the depth map from the network by two ($\frac{H}{2} \times \frac{W}{2}$) before the depth map refinement. The \textit{f\_scores} of \textit{intermediate} and \textit{advanced} sets increase from $43.48$ to $48.40$ and from $24.91$ to $29.55$ respectively. 

\vspace{-4mm}
\paragraph{ETH3D Benchmark \cite{schoeps2017cvpr}}
We also evaluate our method on the recent ETH3D benchmark. The dataset is divided into the \textit{low-res} and the \textit{high-res} scenes, and provides the ground truth depth maps for MVS training. We first fine-tune the model on the ETH3D \textit{low-res} training set, however, observe no performance gain compared to the model only pre-trained on DTU. We suspect the problem may be some images in \textit{low-res} training set are blurred and overexposed as they are captured using hand-held devices. Also, the scenes of ETH3D dataset are complicated in object occlusions, which are not explicitly handled in the proposed network. We evaluate on this benchmark without fine-tuning the network. Our method achieves similar performance to MVSNet \cite{yao2018mvsnet} and ranks $6^{th}$ on the \textit{low-res} benchmark. 

\begin{table}[]
\resizebox{0.48\textwidth}{!}{
\begin{tabular}{c c c c}
\specialrule{.12em}{.06em}{.06em}
                						& Mean Acc. 	& Mean Comp. 	& Overall ($mm$) \\ \hline
Camp \cite{campbell2008using}          	& 0.835      	& 0.554      	& 0.695   \\ 
Furu \cite{furukawa2010accurate}       	& 0.613     	& 0.941      	& 0.777   \\ 
Tola \cite{tola2012efficient}          	& 0.342     	& 1.19       	& 0.766   \\ 
Gipuma \cite{galliani2015massively}    	& \textbf{0.283}& 0.873     	& 0.578   \\ 
Colmap \cite{galliani2015massively}    	& 0.400 		& 0.664     	& 0.532   \\ 
SurfaceNet \cite{ji2017surfacenet}     	& 0.450      	& 1.04       	& 0.745   \\ 
MVSNet (D=256) \cite{yao2018mvsnet} 	& 0.396     	& 0.527      	& 0.462   \\ \hline
R-MVSNet (D=256) 						& 0.385			& 0.459			& 0.422   \\ 
R-MVSNet (D=512) 						& 0.383     	& \textbf{0.452}& \textbf{0.417}   \\ 
\specialrule{.12em}{.06em}{.06em}
\end{tabular}
}
\vspace{-1mm}
\caption{Quantitative results on the DTU evaluation scans \cite{aanaes2016large}. R-MVSNet outperforms all methods in terms of reconstruction completeness and overall quality}
\label{tab:dtu}
\vspace{-2mm}
\end{table}
\begin{figure}[t]
  \centering
  \includegraphics[width=1\linewidth]{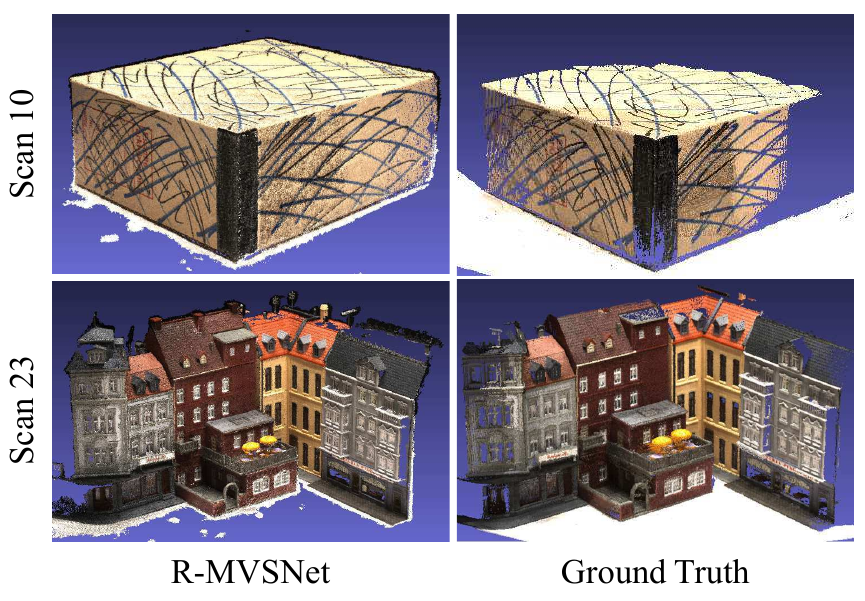}
  \vspace{-6mm}
  \caption{Our results and the ground truth point clouds of scans 10 and 23, DTU \cite{aanaes2016large} dataset} 
  \label{fig:dtu}
  \vspace{-2mm}
\end{figure}

\begin{table*}[]
\resizebox{1\textwidth}{!}{
\begin{tabular}{c|c c c c c c|c c c c c c|c}
\specialrule{.16em}{.08em}{.08em}
\multirow{2}{*}{Dataset} & \multicolumn{6}{c|}{MVSNet\cite{yao2018mvsnet}}   & \multicolumn{6}{c|}{R-MVSNet (Ours)}     & Mem-Util \\ \cline{2-13}
                         			& Rank 	& H & W & Ave. D& Mem.	&Mem-Util & Rank	& H 	& W 	& Ave. D& Mem. &Mem-Util & Ratio	\\ \hline
DTU     \cite{aanaes2016large} 		& 2    	& 1600 	& 1184 	& 256 	& 15.4 GB 	& 1.97 M& 1 & 1600 	& 1200 	& 512 	& 6.7 GB   & 9.17	M& 4.7   \\ 
T. Int. \cite{knapitsch2017tanks}	& 4    	& 1920 	& 1072 	& 256 	& 15.3 GB 	& 2.15 M& 3 & 1920 	& 1080	& 898 	& 6.7 GB   & 17.4	M& 8.1   \\ 
T. Adv. \cite{knapitsch2017tanks}   & -    	& -   	& -   	& -   	& - 		& -		& 3 & 1920 	& 1080 	& 698 	& 6.7 GB   & 13.5	M& - 	\\ 
ETH3D   \cite{schoeps2017cvpr}		& 5    	& 928 	& 480 	& 320 	& 8.7 GB 	& 1.02 M& 6 & 928 	& 480 	& 351 	& 2.1 GB   & 4.65	M& 4.6   \\ 
\specialrule{.16em}{.08em}{.08em}
\end{tabular}
}
\vspace{-2mm}
\caption{Comparisons between MVSNet \cite{yao2018mvsnet} and the proposed R-MVSNet on benchmarking rankings, reconstruction scales and GPU memory requirements on the three MVS datasets \cite{aanaes2016large,knapitsch2017tanks,schoeps2017cvpr}. The memory utility (Mem-Util) measures the data size processed per memory unit, and the high ratio between the two algorithms reflects the scalability of R-MVSNet}
\label{tab:scalability}
\end{table*}

\subsection{Scalability} \label{sec:large}
\noindent
Next, we demonstrate the scalability of R-MVSNet from: 1) wide-range and 2) high-resolution depth reconstructions. 

\vspace{-4mm}
\paragraph{Wide-range Depth Reconstructions} 
The memory requirement of R-MVSNet is independent to the depth sample number $D$, which enables the network to infer depth maps with large depth range that is unable to be recovered by previous learning-based MVS methods. Some large scale reconstructions of Tanks and Temples dataset are shown in Fig.~\ref{fig:tanksandtemples}. Table~\ref{tab:scalability} compares MVSNet \cite{yao2018mvsnet} and R-MVSNet in terms of benchmarking rankings, reconstruction scales and memory requirements. We define the algorithm's memory utility (Mem-Util) as the size of volume processed per memory unit ($\frac{H}{4} \times \frac{W}{4} \times D$ / runtime memory size). R-MVSNet is $\times8$ more efficient than MVSNet in Mem-Util. 

\vspace{-4mm}
\paragraph{High-resolution Depth Reconstructions} 
R-MVSNet can also produce high-resolution depth reconstructions by sampling denser in depth direction. For the DTU evaluation in Sec.~\ref{sec:benchmarks}, if we fix the depth range and change the depth sample number from $D=256$ to $D=512$, the \textit{overall} distance score will be reduced from $0.422mm$ to $0.419mm$ (see last row of Table~\ref{tab:dtu}).

\begin{figure*}
\hspace{-3mm}
\begin{minipage}[h]{.49\textwidth}
    \centering
    \includegraphics[scale=0.75]{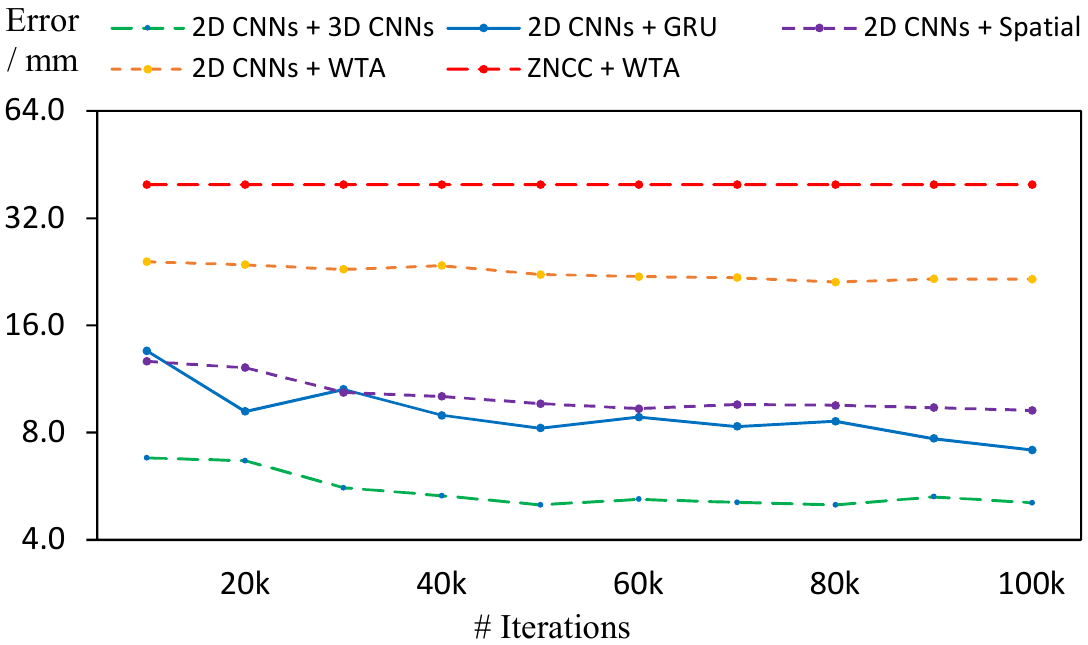}
\end{minipage}
\hspace{3.5mm}
\begin{minipage}[h]{.5\textwidth}
    \centering
	\resizebox{0.96\textwidth}{!}{
	\begin{tabular}{c c c c|c c c}
	\specialrule{.12em}{.06em}{.06em}
     Ref. 	& Pho.	& Geo. 	& Fus.	& Acc. & Comp. & Overall 	\\ \hline
	$\surd$	&$\surd$&$\surd$&$\surd$& \textbf{0.385} & 0.459 & \textbf{0.422}   	\\ 
	{\color{red} $\times$}&$\surd$&$\surd$&$\surd$& 0.444 & 0.486 & 0.465		\\ 
	$\surd$	& {\color{red} $\times$}&$\surd$&$\surd$& 0.550 & \textbf{0.384} & 0.467		\\ 
	$\surd$	&$\surd$&{\color{red} $\times$}& $\surd$	& 0.479 & 0.385 & 0.432		\\ 
	$\surd$	&$\surd$&$\surd$& {\color{red} $\times$}	& 0.498 & 0.364 & 0.431		\\ 
	{\color{red} $\times$}&$\surd$&$\surd$& {\color{red} $\times$}	& 0.605 & 0.373 & 0.489		\\ 
	{\color{red} $\times$}&{\color{red} $\times$}&$\surd$&$\surd$& 0.591 & 0.411 & 0.501		\\ 
	\specialrule{.12em}{.06em}{.06em}
	\end{tabular}
	}
\end{minipage}

\begin{minipage}[t]{.47\textwidth}
    \caption{Ablation studies on network architectures, which demonstrate the importance of learned features and learned regularization. WTA is referred to \textbf{W}inner-\textbf{T}ake-\textbf{A}ll and the figure records testing depth map errors during training}
    \label{fig:ablation_reg}
\end{minipage}
\hspace{7mm}
\begin{minipage}[t]{.47\textwidth}
    \captionof{table}[foo]{Ablation studies on different combinations of variational \textbf{Ref}inement, \textbf{Pho}to-metic filtering, \textbf{Geo}metry filtering and depth map \textbf{Fus}ion for post-processing. Tested on DTU \cite{aanaes2016large} evaluation set}
    \label{fig:ablation_post}
\end{minipage}
\vspace{-5mm}
\end{figure*}
\begin{figure*}[t!]
  \centering
  \includegraphics[width=1\linewidth]{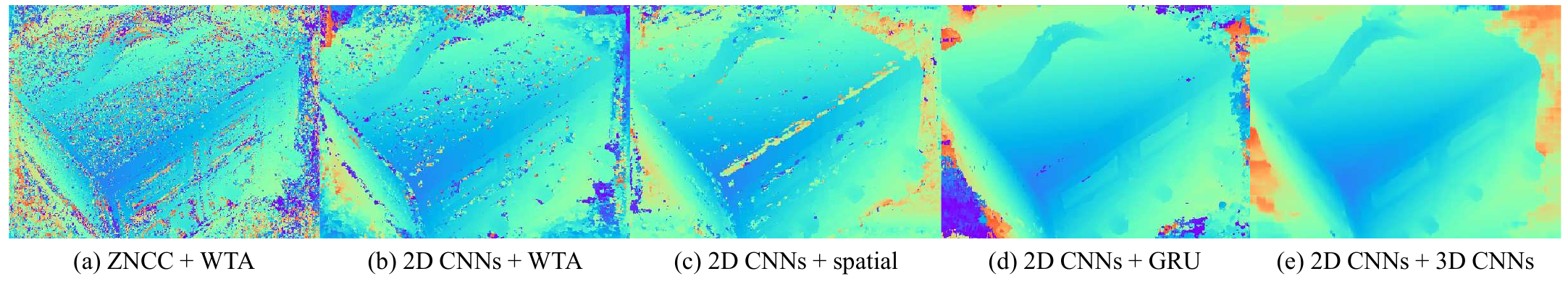}
  \vspace{-5mm}
  \caption{Depth map reconstructions of \textit{scan 11}, DTU dataset \cite{aanaes2016large} using different image features and cost volume regularization methods. All models are trained for 100k iterations}
  \label{fig:compare_reg}
\end{figure*}

\subsection{Ablation studies}

\subsubsection{Networks}
\noindent 
This section studies how different components in the network affect the depth map reconstruction. We perform the study on DTU \textit{validation} set with $W \times H \times D = 640 \times 512 \times 256$, and use the average absolute difference between the inferred and the ground truth depth maps for the quantitative comparison. We denote the learned 2D image features as 2D CNNs. The comparison results of following settings are shown Fig.~\ref{fig:ablation_reg} and Fig.~\ref{fig:compare_reg}:

\vspace{-4mm}
\paragraph{2D CNNs + 3D CNNs} Replace the GRU regularization with the same 3D CNNs regularization in MVSNet \cite{yao2018mvsnet}. As shown in Fig.~\ref{fig:ablation_reg} and Fig.~\ref{fig:compare_reg}, 3D CNNs produces the best depth map reconstructions. 

\vspace{-4mm}
\paragraph{2D CNNs + GRU} The setting of the proposed R-MVSNet, which produces the $2^{nd}$ best depth map results among all settings. The qualitative comparison between 3D CNNs and GRU is shown in Fig.~\ref{fig:compare_reg} (d) and (e).

\vspace{-4mm}
\paragraph{2D CNNs + Spatial} Replace the GRU regularization with the simple spatial regularization. We approach the spatial regularization by a simple 3-layer, 32-channel 2D network on the cost map. The depth map error of spatial regularization is larger than the GRU regularization.

\vspace{-4mm}
\paragraph{2D CNNs + Winner-Take-All} Replace the GRU regularization with simple the winner-take-all selection. We apply a single layer, 1-channel 2D CNN to directly map the cost map to the regularized cost map. The depth map error is further larger than the spatial regularization.

\vspace{-4mm}
\paragraph{ZNCC + Winner-Take-All} Replace the learned image feature and cost metric with the engineered $ZNCC$ (window size of $7 \times 7$). This setting is also referred to the classical plane sweeping \cite{collins1996space}. As expected, plane sweeping produces the highest depth map error among all methods.

\subsubsection{Post-processing}
Next, we study the influences of post processing steps on the final point cloud reconstruction. We reconstruct the DTU \textit{evaluation} without the variational refinement, photo-metric filtering, geometric filtering or depth map fusion. Quantitative results are shown in Table~\ref{fig:ablation_post}.

\vspace{-4mm}
\paragraph{Without Variational Refinement} This setting is similar to the post-processing of MVSNet \cite{yao2018mvsnet}. The \textit{f\_score} is changed to a larger number of 0.465, demonstrating the effectiveness of the proposed depth map refinement.

\vspace{-4mm}
\paragraph{Without Photo-metric Filtering} Table~\ref{fig:ablation_post} shows that the \textit{f\_score} without photo-metric filtering is increased to a larger number of 0.467, which demonstrates the importance of the probability map for photo-metric filtering (Fig.~\ref{fig:pipeline} (f)).

\vspace{-4mm}
\paragraph{Without Geo-metric Filtering} The \textit{f\_score} is increased to 0.432, showing the effectiveness of depth consistency.

\vspace{-4mm}
\paragraph{Without Depth Map Fusion} The \textit{f\_score} is also increased to 0.431, showing the effectiveness of depth fusion.

\subsection{Discussion}

\paragraph{Running Time}
For DTU evaluation with $D=256$, R-MVSNet generates the depth map at a speed of $9.1 s$ / view. Specifically, it takes $2.9 s$ to infer the initial depth map and $6.2 s$ to perform the depth map refinement. It is noteworthy that the runtime of depth map refinement only relates to refinement iterations and the input image size. Filtering and fusion takes neglectable runtime. 

\vspace{-4mm}
\paragraph{Generalization}
R-MVSNet is trained with fixed input size of $N \times W \times H \times D = 3 \times 640 \times 512 \times 256$, but it is applicable to arbitrary input size during testing. It is noteworthy that we use the model trained on the DTU dataset \cite{aanaes2016large} for all our experiments without fine-tuning. While R-MVSNet has shown satisfying generalizability to the other two datasets \cite{knapitsch2017tanks,schoeps2017cvpr}, we hope to train R-MVSNet on a more diverse MVS dataset, and expect better performances on Tanks and Temples \cite{knapitsch2017tanks} and ETH3D \cite{schoeps2017cvpr} benchmarks in the future. 

\vspace{-4mm}
\paragraph{Limitation on Image Resolution}
While R-MVSNet is applicable to reconstructions with unlimited depth-wise resolution, the reconstruction scale is still restricted to the input image size. Currently R-MVSNet can handle a maximum input image size of $3072 \times 2048$ on a 11GB GPU, which covers all modern MVS benchmarks except for the ETH3D high-res benchmark ($6000 \times 4000$).

\section{Conclusions}
\noindent
We presented a scalable deep architecture for high-resolution multi-view stereo reconstruction. Instead of using 3D CNNs, the proposed R-MVSNet sequentially regularizes the cost volume through the depth direction with the convolutional GRU, which dramatically reduces the memory requirement for learning-based MVS reconstructions. 
Experiments show that with the proposed post-processing, R-MVSNet is able to produce high quality benchmarking results as the original MVSNet \cite{yao2018mvsnet}. Also, R-MVSNet is applicable to large-scale reconstructions which cannot be handled by the previous learning-based MVS approaches.

{\small
\bibliographystyle{ieee}
\bibliography{egbib}

\begin{thebibliography}{10}\itemsep=-1pt

\bibitem{aanaes2016large}
H.~Aan{\ae}s, R.~R. Jensen, G.~Vogiatzis, E.~Tola, and A.~B. Dahl.
\newblock Large-scale data for multiple-view stereopsis.
\newblock {\em International Journal of Computer Vision (IJCV)}, 2016.

\bibitem{tensorflow2015-whitepaper}
M.~Abadi, A.~Agarwal, P.~Barham, E.~Brevdo, Z.~Chen, C.~Citro, G.~S. Corrado,
  A.~Davis, J.~Dean, M.~Devin, S.~Ghemawat, I.~Goodfellow, A.~Harp, G.~Irving,
  M.~Isard, Y.~Jia, R.~Jozefowicz, L.~Kaiser, M.~Kudlur, J.~Levenberg,
  D.~Man\'{e}, R.~Monga, S.~Moore, D.~Murray, C.~Olah, M.~Schuster, J.~Shlens,
  B.~Steiner, I.~Sutskever, K.~Talwar, P.~Tucker, V.~Vanhoucke, V.~Vasudevan,
  F.~Vi\'{e}gas, O.~Vinyals, P.~Warden, M.~Wattenberg, M.~Wicke, Y.~Yu, and
  X.~Zheng.
\newblock {TensorFlow}: Large-scale machine learning on heterogeneous systems,
  2015.
\newblock Software available from tensorflow.org.

\bibitem{ballas2015delving}
N.~Ballas, L.~Yao, C.~Pal, and A.~Courville.
\newblock Delving deeper into convolutional networks for learning video
  representations.
\newblock {\em International Conference on Learning Representations (ICLR)},
  2016.

\bibitem{campbell2008using}
N.~D. Campbell, G.~Vogiatzis, C.~Hern{\'a}ndez, and R.~Cipolla.
\newblock Using multiple hypotheses to improve depth-maps for multi-view
  stereo.
\newblock {\em European Conference on Computer Vision (ECCV)}, 2008.

\bibitem{chang2018pyramid}
J.-R. Chang and Y.-S. Chen.
\newblock Pyramid stereo matching network.
\newblock {\em Computer Vision and Pattern Recognition (CVPR)}, 2018.

\bibitem{cho2014learning}
K.~Cho, B.~van Merrienboer, C.~Gulcehre, D.~Bahdanau, F.~Bougares, H.~Schwenk,
  and Y.~Bengio.
\newblock Learning phrase representations using rnn encoder--decoder for
  statistical machine translation.
\newblock {\em Empirical Methods in Natural Language Processing (EMNLP)}, 2014.

\bibitem{collins1996space}
R.~T. Collins.
\newblock A space-sweep approach to true multi-image matching.
\newblock {\em Computer Vision and Pattern Recognition (CVPR)}, 1996.

\bibitem{elman1990finding}
J.~L. Elman.
\newblock Finding structure in time.
\newblock {\em Cognitive science}, 1990.

\bibitem{furukawa2010accurate}
Y.~Furukawa and J.~Ponce.
\newblock Accurate, dense, and robust multiview stereopsis.
\newblock {\em IEEE Transactions on Pattern Analysis and Machine Intelligence
  (TPAMI)}, 2010.

\bibitem{galliani2015massively}
S.~Galliani, K.~Lasinger, and K.~Schindler.
\newblock Massively parallel multiview stereopsis by surface normal diffusion.
\newblock {\em International Conference on Computer Vision (ICCV)}, 2015.

\bibitem{hartmann2017learned}
W.~Hartmann, S.~Galliani, M.~Havlena, L.~Van~Gool, and K.~Schindler.
\newblock Learned multi-patch similarity.
\newblock {\em International Conference on Computer Vision (ICCV)}, 2017.

\bibitem{hirschmuller2008stereo}
H.~Hirschmuller.
\newblock Stereo processing by semiglobal matching and mutual information.
\newblock {\em IEEE Transactions on Pattern Analysis and Machine Intelligence
  (TPAMI)}, 2008.

\bibitem{huang2018deepmvs}
P.-H. Huang, K.~Matzen, J.~Kopf, N.~Ahuja, and J.-B. Huang.
\newblock Deepmvs: Learning multi-view stereopsis.
\newblock {\em Computer Vision and Pattern Recognition (CVPR)}, 2018.

\bibitem{ji2017surfacenet}
M.~Ji, J.~Gall, H.~Zheng, Y.~Liu, and L.~Fang.
\newblock Surfacenet: An end-to-end 3d neural network for multiview stereopsis.
\newblock {\em International Conference on Computer Vision (ICCV)}, 2017.

\bibitem{kar2017learning}
A.~Kar, C.~H{\"a}ne, and J.~Malik.
\newblock Learning a multi-view stereo machine.
\newblock {\em Advances in Neural Information Processing Systems (NIPS)}, 2017.

\bibitem{kendall2017end}
A.~Kendall, H.~Martirosyan, S.~Dasgupta, and P.~Henry.
\newblock End-to-end learning of geometry and context for deep stereo
  regression.
\newblock {\em Computer Vision and Pattern Recognition (CVPR)}, 2017.

\bibitem{knapitsch2017tanks}
A.~Knapitsch, J.~Park, Q.-Y. Zhou, and V.~Koltun.
\newblock Tanks and temples: Benchmarking large-scale scene reconstruction.
\newblock {\em ACM Transactions on Graphics (TOG)}, 2017.

\bibitem{kuhn2017tv}
A.~Kuhn, H.~Hirschm{\"u}ller, D.~Scharstein, and H.~Mayer.
\newblock A tv prior for high-quality scalable multi-view stereo
  reconstruction.
\newblock {\em International Journal of Computer Vision (IJCV)}, 2017.

\bibitem{lhuillier2005quasi}
M.~Lhuillier and L.~Quan.
\newblock A quasi-dense approach to surface reconstruction from uncalibrated
  images.
\newblock {\em IEEE Transactions on Pattern Analysis and Machine Intelligence
  (TPAMI)}, 2005.

\bibitem{li2016efficient}
S.~Li, S.~Y. Siu, T.~Fang, and L.~Quan.
\newblock Efficient multi-view surface refinement with adaptive resolution
  control.
\newblock {\em European Conference on Computer Vision (ECCV)}, 2016.

\bibitem{luo2016efficient}
W.~Luo, A.~G. Schwing, and R.~Urtasun.
\newblock Efficient deep learning for stereo matching.
\newblock {\em Computer Vision and Pattern Recognition (CVPR)}, 2016.

\bibitem{merrell2007real}
P.~Merrell, A.~Akbarzadeh, L.~Wang, P.~Mordohai, J.-M. Frahm, R.~Yang,
  D.~Nist{\'e}r, and M.~Pollefeys.
\newblock Real-time visibility-based fusion of depth maps.
\newblock {\em International Conference on Computer Vision (ICCV)}, 2007.

\bibitem{openMVG}
P.~Moulon, P.~Monasse, R.~Marlet, and Others.
\newblock Openmvg. an open multiple view geometry library.
\newblock \url{https://github.com/openMVG/openMVG}.

\bibitem{riegler2017octnet}
G.~Riegler, A.~O. Ulusoy, and A.~Geiger.
\newblock Octnet: Learning deep 3d representations at high resolutions.
\newblock {\em Computer Vision and Pattern Recognition (CVPR)}, 2017.

\bibitem{schonberger2016pixelwise}
J.~L. Sch{\"o}nberger, E.~Zheng, J.-M. Frahm, and M.~Pollefeys.
\newblock Pixelwise view selection for unstructured multi-view stereo.
\newblock {\em European Conference on Computer Vision (ECCV)}, 2016.

\bibitem{schoeps2017cvpr}
T.~Sch\"ops, J.~L. Sch\"onberger, S.~Galliani, T.~Sattler, K.~Schindler,
  M.~Pollefeys, and A.~Geiger.
\newblock A multi-view stereo benchmark with high-resolution images and
  multi-camera videos.
\newblock {\em Computer Vision and Pattern Recognition (CVPR)}, 2017.

\bibitem{tola2012efficient}
E.~Tola, C.~Strecha, and P.~Fua.
\newblock Efficient large-scale multi-view stereo for ultra high-resolution
  image sets.
\newblock {\em Machine Vision and Applications (MVA)}, 2012.

\bibitem{wang2017cnn}
P.-S. Wang, Y.~Liu, Y.-X. Guo, C.-Y. Sun, and X.~Tong.
\newblock O-cnn: Octree-based convolutional neural networks for 3d shape
  analysis.
\newblock {\em ACM Transactions on Graphics (TOG)}, 2017.

\bibitem{yang2012non}
Q.~Yang.
\newblock A non-local cost aggregation method for stereo matching.
\newblock {\em Computer Vision and Pattern Recognition (CVPR)}, 2012.

\bibitem{yao2018mvsnet}
Y.~Yao, Z.~Luo, S.~Li, T.~Fang, and L.~Quan.
\newblock Mvsnet: Depth inference for unstructured multi-view stereo.
\newblock {\em European Conference on Computer Vision (ECCV)}, 2018.

\bibitem{yoon2006adaptive}
K.-J. Yoon and I.~S. Kweon.
\newblock Adaptive support-weight approach for correspondence search.
\newblock {\em IEEE Transactions on Pattern Analysis and Machine Intelligence
  (TPAMI)}, 2006.

\bibitem{zbontar2016stereo}
J.~Zbontar and Y.~LeCun.
\newblock Stereo matching by training a convolutional neural network to compare
  image patches.
\newblock {\em Journal of Machine Learning Research (JMLR)}, 2016.

\bibitem{zhang2015joint}
R.~Zhang, S.~Li, T.~Fang, S.~Zhu, and L.~Quan.
\newblock Joint camera clustering and surface segmentation for large-scale
  multi-view stereo.
\newblock {\em International Conference on Computer Vision (ICCV)}, 2015.

\bibitem{zhu2018towards}
X.~Zhu, J.~Dai, X.~Zhu, Y.~Wei, and L.~Yuan.
\newblock Towards high performance video object detection for mobiles.
\newblock {\em arXiv preprint arXiv 1804.05830}, 2018.

\end{thebibliography}
}

\newpage

\twocolumn[
\begin{center}
\textbf{\LARGE Supplemental Materials}
\vspace{5mm}
\end{center}
]
\setcounter{section}{0}
\setcounter{equation}{0}
\setcounter{figure}{0}
\setcounter{table}{0}
\setcounter{page}{1}
\makeatletter

\section{Network Architecture}
\noindent
This section describes the network architecture of R-MVSNet (Table~\ref{tab:architecture}). R-MVSNet constructs cost maps at different depths, and recurrently regularizes cost maps through the depth direction. The probability volume need to be explicitly computed during the network training, but for testing, we can sequentially retrieve the regularized cost maps and all layers only require the GPU memory with size linear to the input image resolution.

\begin{table}[h]
\resizebox{.48\textwidth}{!}{%
\begin{tabular}{c|l|c|c}
\specialrule{.2em}{.1em}{.1em}
Output          & Layer  & Input   & Output Size               \\ \hline
$\{\textbf{I}_i\}_{i=1}^N$ &                                    &        & N$\times$H$\times$W$\times$3\\ \hline
\multicolumn{4}{c}{\textbf{Image Features Extration}}                                                        \\ \hline
2D\_0           & ConvBR,K=3x3,S=1,F=8      & $\mathbf{I}_i$  & H$\times$W$\times$8                  \\ 
2D\_1           & ConvBR,K=3x3,S=1,F=8      & 2D\_0           & H$\times$W$\times$ 8                  \\ 
2D\_2           & ConvBR,K=5x5,S=2,F=16     & 2D\_1           & \sfrac{1}{2}H$\times$\sfrac{1}{2}W$\times$16         \\ 
2D\_3           & ConvBR,K=3x3,S=1,F=16     & 2D\_2           & \sfrac{1}{2}H$\times$\sfrac{1}{2}W$\times$16         \\ 
2D\_4           & ConvBR,K=3x3,S=1,F=16     & 2D\_3           & \sfrac{1}{2}H$\times$\sfrac{1}{2}W$\times$16         \\ 
2D\_5           & ConvBR,K=5x5,S=2,F=32     & 2D\_4           & \sfrac{1}{4}H$\times$\sfrac{1}{4}W$\times$32         \\ 
2D\_6           & ConvBR,K=3x3,S=1,F=32     & 2D\_5           & \sfrac{1}{4}H$\times$\sfrac{1}{4}W$\times$32         \\ 
$\mathbf{F}_i$  & Conv,K=3x3,S=1,F=32        & 2D\_6           & \sfrac{1}{4}H$\times$\sfrac{1}{4}W$\times$32      \\ \hline
\multicolumn{4}{c}{\textbf{Differentiable Homography Warping}}                                                      \\ \hline
$\{\textbf{F}_i,\textbf{H}_i(d)\}_{i=1}^N$ & DH-Warping & $\{\textbf{V}_i(d)\}_{i=1}^N$ & \sfrac{1}{4}H$\times$\sfrac{1}{4}W$\times$32 \\ \hline
\multicolumn{4}{c}{\textbf{Cost Map Construction}}                                                                           \\ \hline
$\{\textbf{V}_i(d)\}_{i=1}^N$ & Variance Cost Metric & $\textbf{C}_0(d)$ & \sfrac{1}{4}H$\times$\sfrac{1}{4}W$\times$32 \\ \hline
\multicolumn{4}{c}{\textbf{GRU Regularization}}                                                                     \\ \hline
$\mathbf{C}(d)$ 					& Conv,K=3x3,S=1,F=16  & $\mathbf{C}_0(d)$ & \sfrac{1}{4}H$\times$\sfrac{1}{4}W$\times$16 \\ \hline
$\mathbf{C}_0(d)\&\mathbf{C}_1(d-1)$& GRU, K=3x3, F=16   & $\mathbf{C}_1(d)$ & \sfrac{1}{4}H$\times$\sfrac{1}{4}W$\times$16 \\\hline
$\mathbf{C}_1(d)\&\mathbf{C}_2(d-1)$& GRU, K=3x3, F=4   & $\mathbf{C}_2(d)$ & \sfrac{1}{4}H$\times$\sfrac{1}{4}W$\times$4 \\\hline
$\mathbf{C}_2(d)\&\mathbf{C}_r(d-1)$& GRU, K=3x3, F=1   & $\mathbf{C}_r(d)$ & \sfrac{1}{4}H$\times$\sfrac{1}{4}W$\times$1 \\\hline
\multicolumn{4}{c}{\textbf{Probability Volume Construction}}                                                                           \\ \hline
$\{\textbf{C}_r(d)\}_{d=1}^D$ & Softmax & $\{\textbf{P}_r(d)\}_{d=1}^N$ & \sfrac{1}{4}H$\times$\sfrac{1}{4}W$\times$D \\ 
\specialrule{.2em}{.1em}{.1em}
\end{tabular}%
}
\centering
\caption{R-MVSNet architecture. We denote the 2D convolution as Conv and use BR to abbreviate the batch normalization and the Relu. K is the kernel size, S the kernel stride and F the output channel number. N, H, W, D denote input view number, image width, height and depth sample number respectively}
\label{tab:architecture}
\end{table}

\section{Depth Sample Number}
\noindent
Given the depth range $[d_{min}, d_{max}]$, we sample depth values using the inverse depth setting:
\begin{equation}\label{eq:inverse}
d(i) = \big( (\frac{1}{d_{min}} - \frac{1}{d_{max}}) \frac{i}{D - 1} + \frac{1}{d_{max}} \big)^{-1}, i \in [1, D]
\end{equation}
where $i$ is the index of the depth sampling and $D$ is the depth sample number. To determine the sample number $D$, we assume that the spatial image resolution should be the same as the temporal depth resolution. Supposing $\mathbf{X}_1$ and $\mathbf{X}_2$ are two 3D points by projecting the reference image center $(\frac{W}{2}, \frac{H}{2})$ and its neighboring pixel $(\frac{W}{2} + 1, \frac{H}{2})$ to the space at depth $d_{min}$, the spatial image resolution at depth $d_{min}$ is defined as $\rho = ||\mathbf{X}_2 - \mathbf{X}_1||_2$. Meanwhile, we define the temporal depth resolution at depth $d_{min}$ as $d(2) - d(1)$. Considering Equation~\ref{eq:inverse}, the depth sample number is calculated as:
\begin{equation}
D = (\frac{1}{d_{min}} - \frac{1}{d_{max}}) / (\frac{1}{d_{min}} - \frac{1}{d_{min} + \rho}).
\end{equation}

\begin{figure*}
  \centering
  \includegraphics[width=1\linewidth]{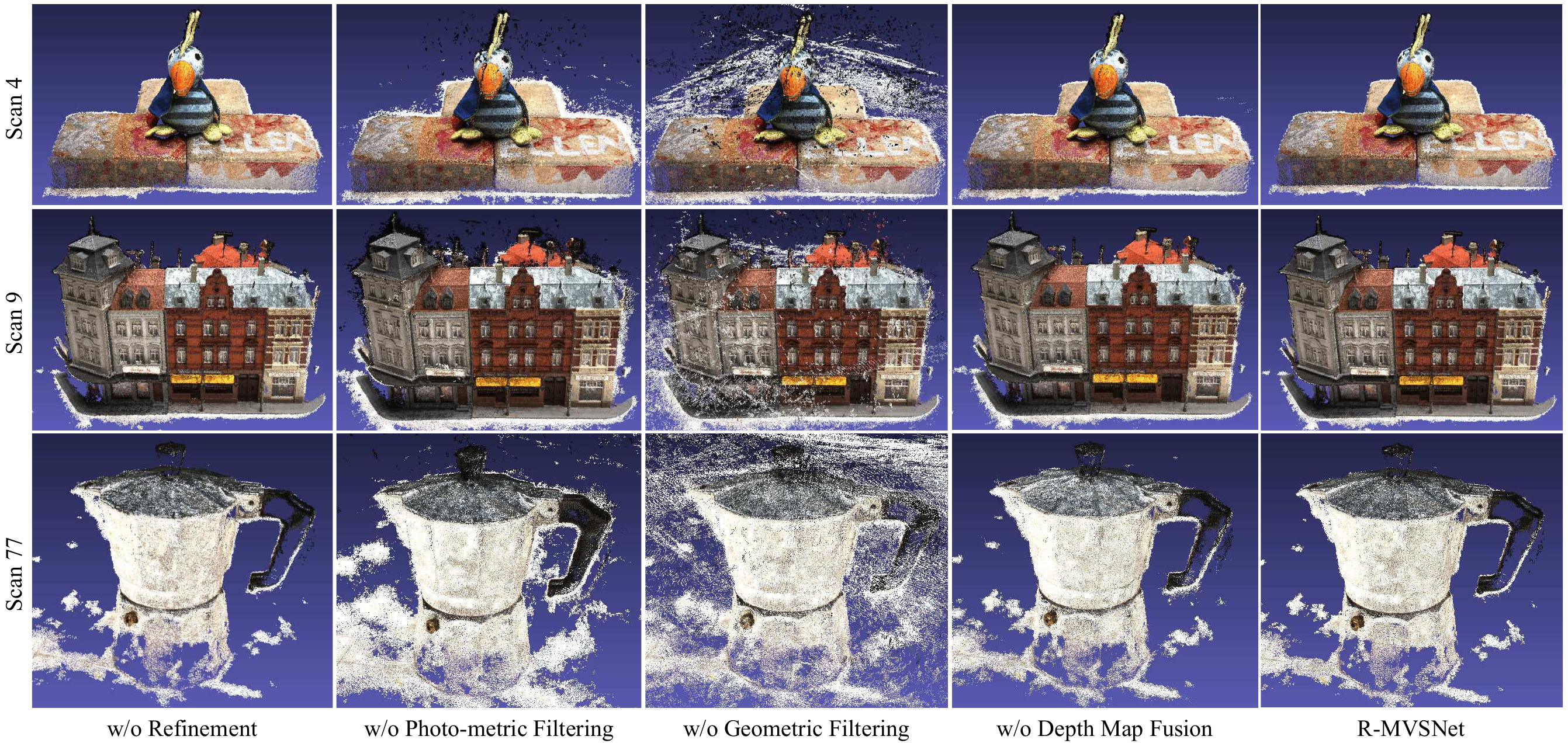}
  \vspace{-5mm}
  \caption{Point cloud reconstructions of DTU dataset~\cite{aanaes2016large} with different post-processing settings}
  \label{fig:ablation}
\end{figure*}

\section{Variational Depth Map Refinement}
\noindent
We derive the iterative minimization procedure for Equation 8 in the main paper. Focusing on one pixel $\mathbf{p}_1$ in the reference image, we denote its corresponding 3D point in the space as $\mathbf{X} = \mathbf{\Pi}_1^{-1}(\mathbf{p}_1) \cdot d_1 + \mathbf{c}_1$, where $\mathbf{\Pi}_1$, $\mathbf{c}_1$ and $d_1$ are the projection matrix, camera center of the reference camera and the depth of pixel $\mathbf{p_1}$. The projection of $\mathbf{X}$ in the source image is $\mathbf{p}_i = \mathbf{\Pi}_i (\mathbf{X})$. For the photo-consistency term, we assume $\mathcal{C}(\mathbf{I}_1(\mathbf{p_1}), \mathbf{I}_{i \to 1}(\mathbf{p}_1)) = \mathcal{C}(\mathbf{I}_{1 \to i}(\mathbf{p}_i), \mathbf{I}_i(\mathbf{p}_i))$ and abbreviate it as $\mathcal{C}_{1 \to i}(\mathbf{p}_i)$. The image reprojection error will be changed as $\mathbf{D}_1$ deforms, and we take the derivative of the photo-consistency term w.r.t. to depth $d_1$:
\begin{equation}
\begin{split}
\frac{\partial E^i_{photo}(\mathbf{p}_1)}{\partial d_1} &= \frac{\partial \mathcal{C}_{1 \to i}(\mathbf{p}_i)}{\partial d_1} \\
&= \frac{\partial \mathcal{C}_{1 \to i}(\mathbf{\Pi}_i (\mathbf{\Pi}_1^{-1}(\mathbf{p}_1) \cdot d_1 + \mathbf{c}_1)}{\partial d_1} \\
&= \frac{\partial \mathcal{C}_{1 \to i}(\mathbf{p}_i)}{\partial \mathbf{p}_i} \cdot \frac{{\partial \mathbf{p}_i}}{\partial \mathbf{X}} \cdot \frac{{\partial \mathbf{X}}}{\partial d_1} \\
&= \frac{\partial \mathcal{C}_{1 \to i}(\mathbf{p}_i)}{\partial \mathbf{p}_i} \cdot \mathbf{J}_i \cdot \mathbf{\Pi}_1^{-1}(\mathbf{p}_1)
\end{split}
\end{equation}
where $\mathbf{J}_i$ is the Jacobian of the projection matrix $\mathbf{\Pi}_i$. $\frac{\partial \mathcal{C}_{1 \to i}(\mathbf{p}_i)}{\partial \mathbf{p}_i}$ is the derivative of the photo-metric measurement w.r.t. the pixel coordinate. For computing the derivatives of NCC and ZNCC, we refer readers to \cite{li2016efficient} for detailed implementations. Also, considering $d_1 = \mathbf{D}_1(\mathbf{p_1})$, the derivative of the smoothness term $\mathcal{S}(\mathbf{p},\mathbf{p'}) = w(\mathbf{p}_1, \mathbf{p}_1') (\mathbf{D}_1(\mathbf{p}) - \mathbf{D}_1(\mathbf{p}'))^2$ can be derived as:
\begin{equation}
\begin{split}
&\frac{\partial E^i_{smooth}(\mathbf{p}_1)}{\partial d_1}  \\
&\quad= \sum_{\mathbf{p}_1' \in \mathcal{N}(\mathbf{p}_1)}w(\mathbf{p}_1, \mathbf{p}_1') \frac{\partial (\mathbf{D}_1(\mathbf{p}_1) - \mathbf{D}_1(\mathbf{p}_1'))^2}{\partial d_1}  \\
&\quad= \sum_{\mathbf{p}_1' \in \mathcal{N}(\mathbf{p}_1)}2w(\mathbf{p}_1, \mathbf{p}_1')(\mathbf{D}_1(\mathbf{p}_1) - \mathbf{D}_1(\mathbf{p}_1'))  
\end{split}
\end{equation}
where $w(\mathbf{p}_1, \mathbf{p}_1') = \exp(-\frac{(\mathbf{I}_1(\mathbf{p_1}) - \mathbf{I}_1(\mathbf{p}_1'))^2}{10})$ is the bilateral smoothness weighting.

We iteratively minimize the total image reprojection error $E$ by gradient descent with a descending step size of $\lambda(t) = 0.9 \cdot \lambda(t-1)$ and $\lambda(0) = 10$. The reference depth map $\mathbf{D}_1$ and all reprojected images $\{\mathbf{I}_{1 \to i}\}_{i=2}^{N}$ will be updated at each step. The refinement iteration is fixed to 20 for all our experiments.

\section{Sliding Window 3D CNNs}
\begin{figure}[t!]
  \centering
  \includegraphics[width=1\linewidth]{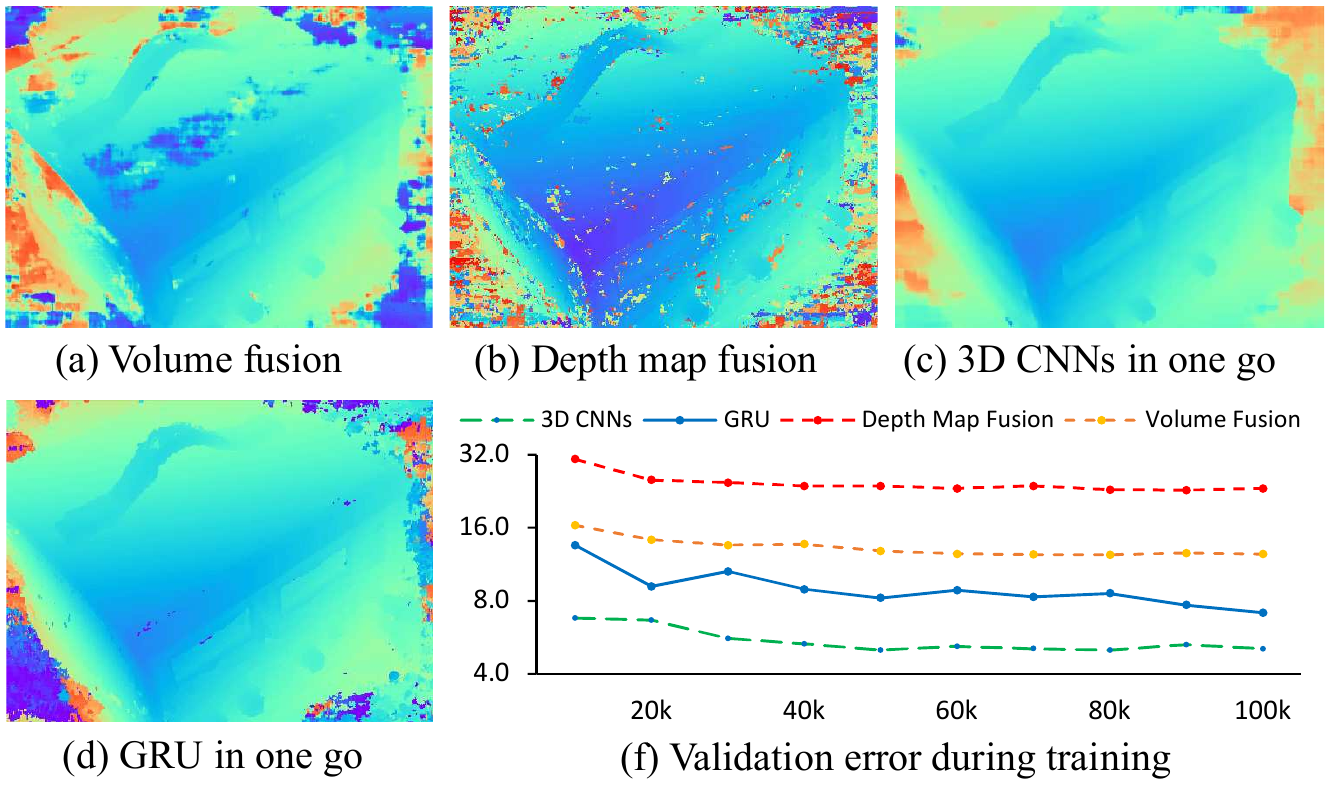}
  \vspace{-2mm}
  \caption{Sliding window 3D CNNs. (a) and (b) are depth map results of the proposed two fusion strategies in A1.}
  \label{fig:sliding}
\end{figure}

\noindent
One concern about R-MVSNet is that whether the proposed GRU regularization could be simply replaced by streaming the 3D CNNs regularization in the depth direction. To address this concern, we conduct two more ablation studies. For DTU dataset, we divide the cost volume $\mathbf{C}$ ($D=256$) into sub-volumes ($D_{sub}=64$) along the depth direction. To better regularize the boundary voxels, we set the overlap between two adjacent sub-volumes to $D_{overlap} = 32$, so in this way $\mathbf{C}$ is divided into 7 subsequent sub-volumes $\{\mathbf{C}_i\}_{i=0}^6$. We then sequentially apply 3D CNNs (except for the softmax layer) on $\{\mathbf{C}_i\}_{i=0}^6$ to obtain the regularized sub-volumes. Then, we generate the final depth map by two different fusion strategies:
\vspace{-1mm}
\begin{itemize}
\item \textbf{Volume Fusion} First concatenate the regularized sub-volumes (truncated with $D_{trunc}=16$ to fit the overlap region) in depth direction. Then apply softmax and soft argmin to regress the final depth map.
\vspace{-1mm}
\item \textbf{Depth Map Fusion} First regress 7 depth maps and probability maps from the regularized sub-volumes. Then fuse the 7 depth maps into the final depth map by winner-take-all selection on probability maps.
\end{itemize}
\vspace{-1mm}
Qualitative and quantitative results are shown in Fig.~\ref{fig:sliding}. Both sliding strategies produce errors higher than GRU and 3D CNNs. Also, sliding strategies take $\sim10s$ to infer depth map ($H \times W \times D = 1600 \times 1184 \times 256$), which is $\sim2\times$ slower than MVSNet and R-MVSNet.

The sliding window 3D CNNs regularization is a depth-wise divide-and-conquer algorithm and there are two major limitations: 1) One is the discrepancies among sub-volumes, as sub-volumes are not regularized as a whole. 2) The second is the limited size of the sub-volume, which is far less than the actual receptive field size of the multi-scale 3D CNNs ($\sim256^3$). As a result, such strategies cannot be fully benefit from the powerful 3D CNNs regularization. 

\section{Post-processing}
\noindent
We show in Fig.~\ref{fig:ablation} the qualitative point cloud results of DTU \textit{evaluation} set~\cite{aanaes2016large} using different post-processing settings. The photo-metric filtering and the geometric filtering are able to remove different kinds of outliers and produce visually clean point clouds. Depth map refinement and depth map fusion have little influence on the qualitative results, however, they are able to reduce the $overall$ score for the quantitative evaluation (Table 3 in the main paper).

\section{Point Cloud Results}
\begin{figure*}[]
  \centering
  \includegraphics[width=1\linewidth]{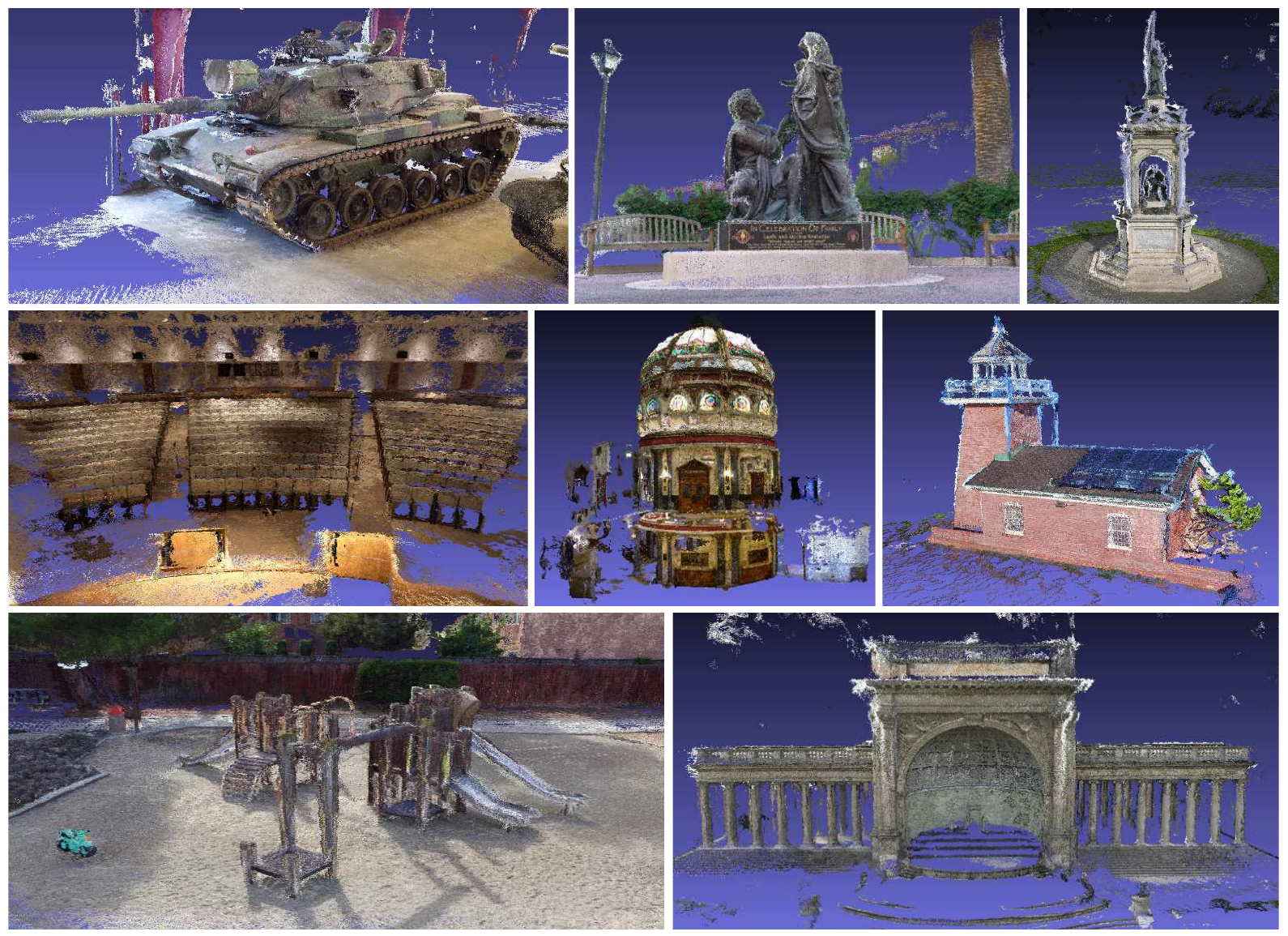}
  \vspace{-5mm}
  \caption{Point cloud reconstructions of Tanks and Temples dataset \cite{knapitsch2017tanks}}
  \label{fig:tt}
\end{figure*}

\begin{figure*}[]
  \centering
  \includegraphics[width=1\linewidth]{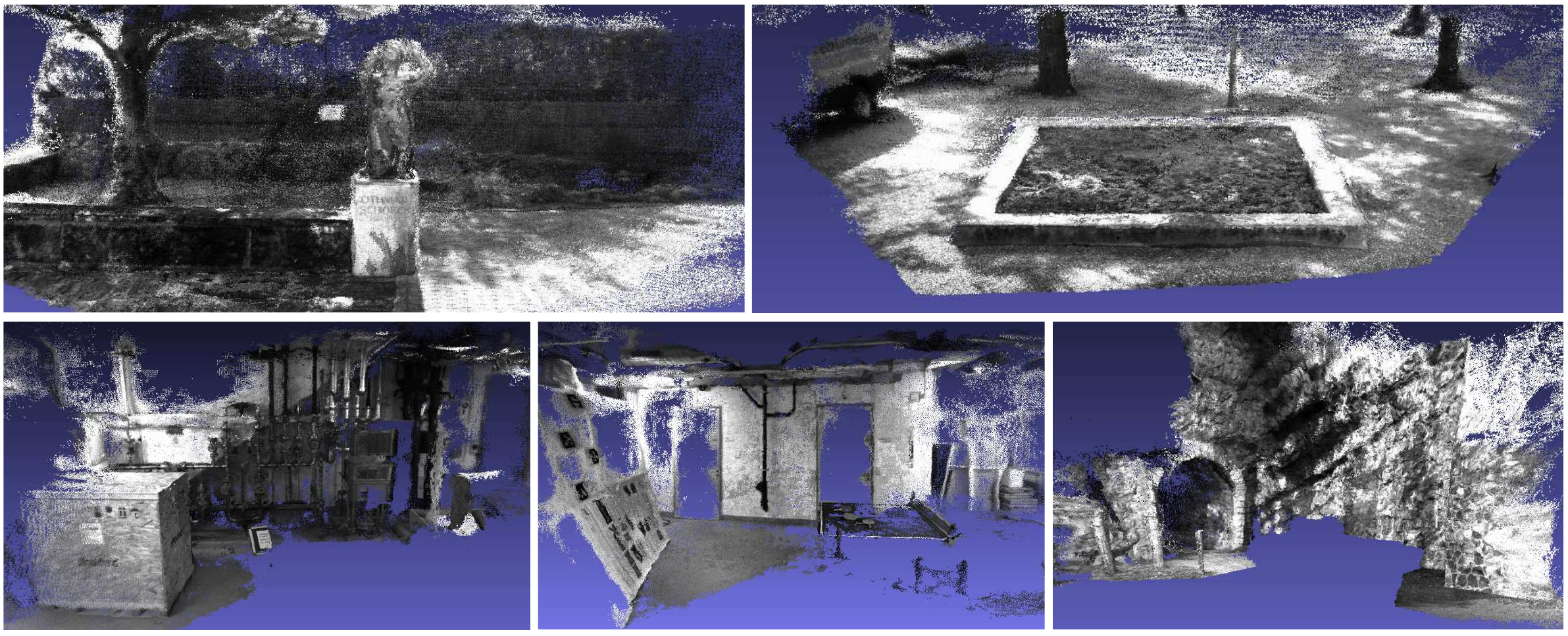}
  \vspace{-5mm}
  \caption{Point cloud reconstructions of ETH3D \textit{low-res} dataset \cite{schoeps2017cvpr}}
  \label{fig:eth3d}
\end{figure*}

\noindent
This section presents the point cloud reconstructions of DTU dataset \cite{aanaes2016large}, Tanks and Temples benchmark \cite{knapitsch2017tanks} and ETH3D benchmark \cite{schoeps2017cvpr} that have not been shown in the main paper. The point cloud results of the three datasets can be found in Fig. \ref{fig:dtu}, Fig. \ref{fig:tt} and Fig. \ref{fig:eth3d} respectively. R-MVSNet is able to produce visually clean and complete point cloud for all reconstructions.

\begin{figure*}[]
  \centering
  \includegraphics[width=1\linewidth]{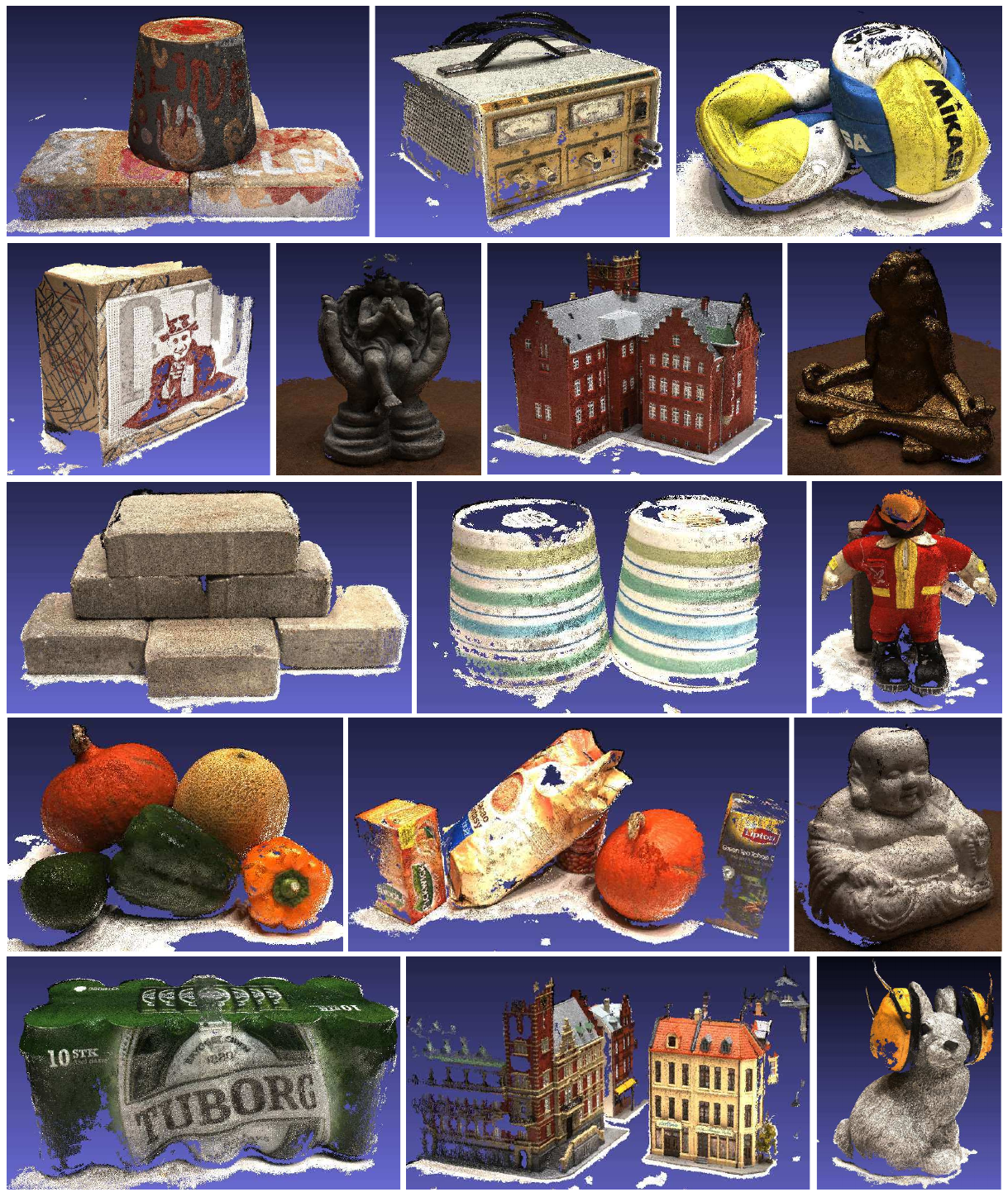}
  \vspace{-5mm}
  \caption{Point cloud reconstructions of DTU \textit{evaluation} set \cite{aanaes2016large}}
  \label{fig:dtu}
\end{figure*}

\end{document}